\documentclass{article}

 \usepackage[final]{neurips_2022}
 \usepackage{graphicx}
\usepackage{subcaption}
\usepackage{float}
\usepackage{hyperref}

\usepackage[utf8]{inputenc} 
\usepackage[T1]{fontenc}    
\usepackage{hyperref}       
\usepackage{url}            
\usepackage{booktabs}       
\usepackage{amsfonts}       
\usepackage{nicefrac}       
\usepackage{microtype}      
\usepackage{xcolor}         

\title{Improving Knowledge Distillation for BERT Models: Loss Functions, Mapping Methods, and Weight Tuning}

\author{%
  Adeem Jassani\\
  MSc in Applied Computing \\
  \texttt{adeemj@cs.toronto.edu} \\
   \And
     Apoorv Dankar\\
  MSc in Applied Computing \\
  \texttt{dankar@cs.toronto.edu} \\
   \And
     Kartikaeya Kumar\\
  MSc in Applied Computing \\
  \texttt{kartikaeya@cs.toronto.edu} \\
}

\begin{document}

\maketitle

\begin{abstract}
  The use of large transformer-based models such as BERT, GPT, and T5 has led to significant advancements in natural language processing. However, these models are computationally expensive, necessitating model compression techniques that reduce their size and complexity while maintaining accuracy. This project investigates and applies knowledge distillation for BERT model compression, specifically focusing on the TinyBERT student model. We explore various techniques to improve knowledge distillation, including experimentation with loss functions, transformer layer mapping methods, and tuning the weights of attention and representation loss and evaluate our proposed techniques on a selection of downstream tasks from the GLUE benchmark. The goal of this work is to improve the efficiency and effectiveness of knowledge distillation, enabling the development of more efficient and accurate models for a range of natural language processing tasks.
\end{abstract}

\section{Introduction}

Large transformer-based models have significantly advanced the field of NLP. However, these models are computationally expensive, posing challenges in deployment on resource-constrained devices. Model compression techniques, like knowledge distillation, aim to reduce model size and complexity while maintaining performance.

This project focuses on improving knowledge distillation for BERT (\cite{devlin2018pretraining}) model compression, specifically using the TinyBERT (\cite{jiao2019tinybert}) student model. We explore various techniques to enhance the efficiency and effectiveness of knowledge distillation, including experimentation with alternative loss functions, transformer layer mapping methods, and tuning the weights of attention and representation loss. We evaluate the performance of our proposed techniques on a selection of downstream tasks from the GLUE benchmark (\cite{wang-etal-2018-glue}). Our objective is to contribute to the development of more efficient and accurate models for a range of natural language processing tasks.

Our experiments (GitHub link in the reproducibility section of Appendix \ref{reproducibility}) show that certain alternative loss functions, such as KL divergence (\cite{kullback1951information}), can lead to improved performance in some scenarios, particularly when training with limited data for challenging tasks like CoLA \cite{warstadt2018neural}. Additionally, we demonstrate the importance of the mapping function between student and teacher layers. Our novel LearnableMap offers a slight improvement over the baseline. Our work highlights the potential benefits of these techniques and provides insights that can inform future research on model compression and knowledge distillation.

\section{Related Works}

Transformer-based models like BERT are commonly compressed using a combination of knowledge distillation, pruning, and quantization techniques. BERT (\cite{devlin2018pretraining}), is one of the most widely used pre-trained language models. Knowledge distillation, introduced by (\cite{hinton2015distilling}), is a popular technique for model compression that involves training a smaller, simpler model to mimic the behavior of a larger, more complex model. DistilBERT (\cite{Sanh2019DistilBERTAD}), a compressed version of BERT, is the first BERT model compressed using knowledge distillation.

To further improve knowledge distillation, researchers have proposed several extensions to the technique. For example, Patient Knowledge Distillation (\cite{Sun2019PatientKD}), uses loss calculated at multiple intermediate layers of the teacher model for incremental knowledge extraction, resulting in better  accuracy. Another notable example is TinyBERT (\cite{jiao2019tinybert}), which refines previous methods by, first, incorporating an attention loss between intermediate layers, and second, utilizing a two-stage distillation process to boost the efficiency and effectiveness of the technique.

\section{Methodology}

In this section, we outline our methodology, detailing the TinyBERT distillation process and the three exploration techniques we employed: alternative loss functions, mapping functions, and tuning the weights of representation and attention loss.

The total loss function for TinyBERT distillation consists of three components: 1) \textbf{Embedding loss} ($\mathcal{L}_{\mathrm{embd}}$) for the input layer ($m=0$), the 2) \textbf{Transformer layer loss} which is a sum of hidden representation loss ($\mathcal{L}_{\mathrm{hidn}}$) and attention loss ($\mathcal{L}_{\mathrm{attn}}$) for the transformer layers ($0<m \le M$), and 3) \textbf{Prediction loss} ($\mathcal{L}_{\mathrm{pred}}$) for the output layer ($m=M+1$). Here, $m$ represents the index of the student layer, and $M$ denotes the total number of transformer layers in the student model.

For the embedding and prediction layers, the loss is computed between the corresponding student and teacher layers. Since there are 12 transformer layers in BERT and 4 transformer layers in TinyBERT, a mapping function $n=g(m)$ (explained in more detail in \ref{sec:mthodology_mapping}) is used to determine the mapped teacher layer for the $m^{th}$ student layer. Thus, for the intermediate layers, the loss is computed between the student layer and the mapped teacher layer. The mapping function and its detailed explanation can be found in Section 3.2.

The TinyBERT distillation process has two stages: 1) \textbf{General Distillation}, which distils the general BERT model into the general TinyBERT model, and 2) \textbf{Task Distillation}, which distils the task-specific BERT model into the task-specific TinyBERT model, initialized from the general TinyBERT model. Task Distillation has two substages: a) Transformer Layer Distillation, using only transformer layer loss, and b) Prediction Layer Distillation, using only prediction layer loss.


\subsection{Exploring KL Divergence as Attention Loss }
 Attention loss, denoted as $\mathcal{L}_{\mathrm {attn }}$, measures the difference between the attention distributions of the teacher and student models. The original TinyBERT approach uses mean squared error (MSE) over unnormalized attention matrices. In our experiments, we use Kullback-Leibler (KL) divergence (\cite{kullback1951information}) over normalized attention matrices as shown in the following equation. 

\begin{equation}
\frac{1}{h} \sum_{i=1}^h M S E\left(A_i^S, A_i^T\right) \rightarrow \frac{1}{h} \sum_{i=1}^h K L\left(\sigma\left(A_i^S\right), \sigma\left(A_i^T\right)\right)
\end{equation}
where $h$ is the number of attention heads, ${A}_i$ refers to the attention matrix corresponding to the $i$-th head of teacher or student and $\sigma$ represents softmax. We explore KL Divergence since it measures the difference between two probability distributions making it a natural fit for comparing attention distributions.

\subsection{Exploring different Mapping Functions}
\label{sec:mthodology_mapping}

As described previously, the mapping functions determine which teacher layers a student layer learns from. In the base paper, a uniform mapping function $g(m) = 3m$ is used. For example, in the uniform mapping, student layer 1 is mapped to teacher layer 3, and student layer 2 is mapped to teacher layer 6, and so on.

Inspired by uniform mapping, we divide the teacher layers into blocks equal to the number of student layers. For example, teacher layers (1, 2, 3) form block 1, (4, 5, 6) form block 2, and so on. We explore various novel mapping functions that combine different teacher layers to create an aggregated representation for the student to learn from. In our framework, we represent the mapping function as the normalized vector ${v}(m) \in \mathbb{R}^3$, as there are three layers in one block. \footnote{We use the value 3 for ease of understanding, but the framework is extensible to any distillation where the number of layers of the teacher is divisible by the number of layers of the student.} The aggregated teacher representation is computed as a linear combination of the representations from the corresponding block with weights ${v}(m)$. The base uniform mapping fits into our framework with ${v}(m) = (0, 0, 1)$. We experiment with the following mapping functions:

\begin{enumerate}
\item \textbf{Random map:} ${v}_k(m) = \mathrm{OneHot}(\mathrm{RandInt}(3)) 
 \forall m$. One layer from the corresponding block is randomly selected to be the teacher representation for the student layer during each iteration. Here, $k$ represents the iteration number during the training process.

\item \textbf{Mean map:} ${v}(m) = \left(\frac{1}{3}, \frac{1}{3}, \frac{1}{3}\right) \forall m$. This (fixed) mapping equally combines all three layers in the corresponding block to form the aggregate teacher representation for the student layer to learn from. 

\item \textbf{Learnable map:} ${v}_k(m) = \mathrm{softmax}({\theta}_k[m]) \forall m$. This mapping learns the optimal combination of layers in the corresponding block to form the aggregate teacher representation. Here, $ \theta \in \mathbb{R}^{(M, N/M)}$ where $M$ and $N$ are numbers of layers in student and teacher respectively.  
\end{enumerate}

\subsection{Tuning the Weights of Representation and Attention Loss}

In the base paper, the authors employ an equal weighting of hidden representation loss and attention loss for transformer layer distillation. In our experiments, we explore adjusting the relative weights of these two losses to examine their impact. We modify the loss function by introducing a tunable parameter $\alpha$ ranging from 0 to 1 (with $\alpha = 0.5$ giving the baseline) , as shown in the modified combined loss equation: 
\begin{equation}
    \mathcal{L}_{\mathrm{hidn}} + \mathcal{L}_{\mathrm{attn}} \rightarrow 2\alpha \cdot \mathcal{L}_{\mathrm{hidn}} + 2(1 - \alpha) \cdot \mathcal{L}_{\mathrm{attn}}  
\end{equation}


\section{Experiments}
In our research project, we focused on the task-specific distillation stage due to the high computational cost associated with general distillation. Thus, we conducted multiple experiments using two datasets from the GLUE benchmark (\cite{wang-etal-2018-glue}) - CoLA and STS-B (\cite{cer2017semeval})  - as they exhibited a significant performance gap between the Teacher and base Student models. The standard metrics for the respective tasks were employed to evaluate the models. The Teacher models used in our experiments were fine-tuned BERT models, which were obtained from the TextAttack Model Hub (\cite{morris2020textattack}) on Hugging Face. We utilized the General TinyBERT model provided by the TinyBERT authors on Hugging Face as our Student model.

Since all our experiments are modifications of the transformer layer loss, we have implemented our experiments only in the transformer distillation sub stage of task distillation stage. We adhered to hyperparameters recommended in the TinyBERT paper (details in \ref{reproducibility}). To keep track of all experiments, we used wandb (\cite{biewald2020experiment}) as our experiment management tool. The experiments were conducted using the NVIDIA GeForce RTX 2060 Rev, which has 6 GB of GDDR6, available from the Department of computer science GPU cluster.

\subsection{Loss Function Experiments}
We conducted experiments to evaluate the performance of the KL divergence loss and compared it with the baseline loss function using full and half training data.

Our experimental results demonstrated that the KL divergence loss outperformed the baseline loss function in the CoLA task. Specifically, we observed a 5.7\% improvement over the baseline when using the full training data and a 12.0\% improvement over the baseline when using half of the training data. However, we did not observe any significant improvement over the baseline for the STS-B task when using the KL divergence loss. The detailed results can be found in the left panels of Figures \ref{fig:cola_results} and \ref{fig:stsb_results}. 

These findings highlight the potential benefits of using the KL divergence loss function in specific scenarios, such as when training with limited data and for challenging tasks like CoLA. 

\subsection{Mapping Functions}
For the mapping function we experimented with the RandomMap, MeanMap and LearnableMap functions as explained in Section 3.2. Additionally in the LearnableMap method, we tried two different initializations of the parameters - ${v}_0(m) = \left(\frac{1}{3}, \frac{1}{3}, \frac{1}{3}\right)$ which is equivalent to MeanMap and ${v}_0(m) = (0.11, 0.11, 0.78)$ 
\footnote{${v}_0(m) = (0.11, 0.11, 0.78)$ is obtained by applying softmax(-1, -1, 1). To obtain the weight vector $(0,0,1)$ we will need to set the  $\theta_0$ set to (0, 0, +inf). However, setting a component of $\theta_0$ to +inf makes the optimization difficult}
which is close to the Base map. We also tuned the independent learning rate of these params among [1e-3, 1e-4, 5e-5, 1e-6]. We found that the baseline was performing significantly better than RandomMap in CoLA which shows the importance of the mapping function. LearnableMap performs comparably to the baseline for CoLA and does not show any major improvement.
However, for the STS-B task all mappings, including random performed almost equally. The detailed results are in the center panels of Figures \ref{fig:cola_results} and \ref{fig:stsb_results}. 

While the learnable mapping doesn't yield significant improvements, an intriguing observation is that the weight vectors ${v}_k(m)$ converge to the same values regardless of initialization (\ref{weights_learnablemap}). This may indicate convergence to a global minimum and the existence of a mapping that minimizes transformer loss, presenting a potentially interesting area for further exploration.

\subsection{Weighting Attention and Representation Loss}
\begin{figure}[h]
  \centering
  \begin{subfigure}{0.28\textwidth}
    \includegraphics[width=\linewidth]{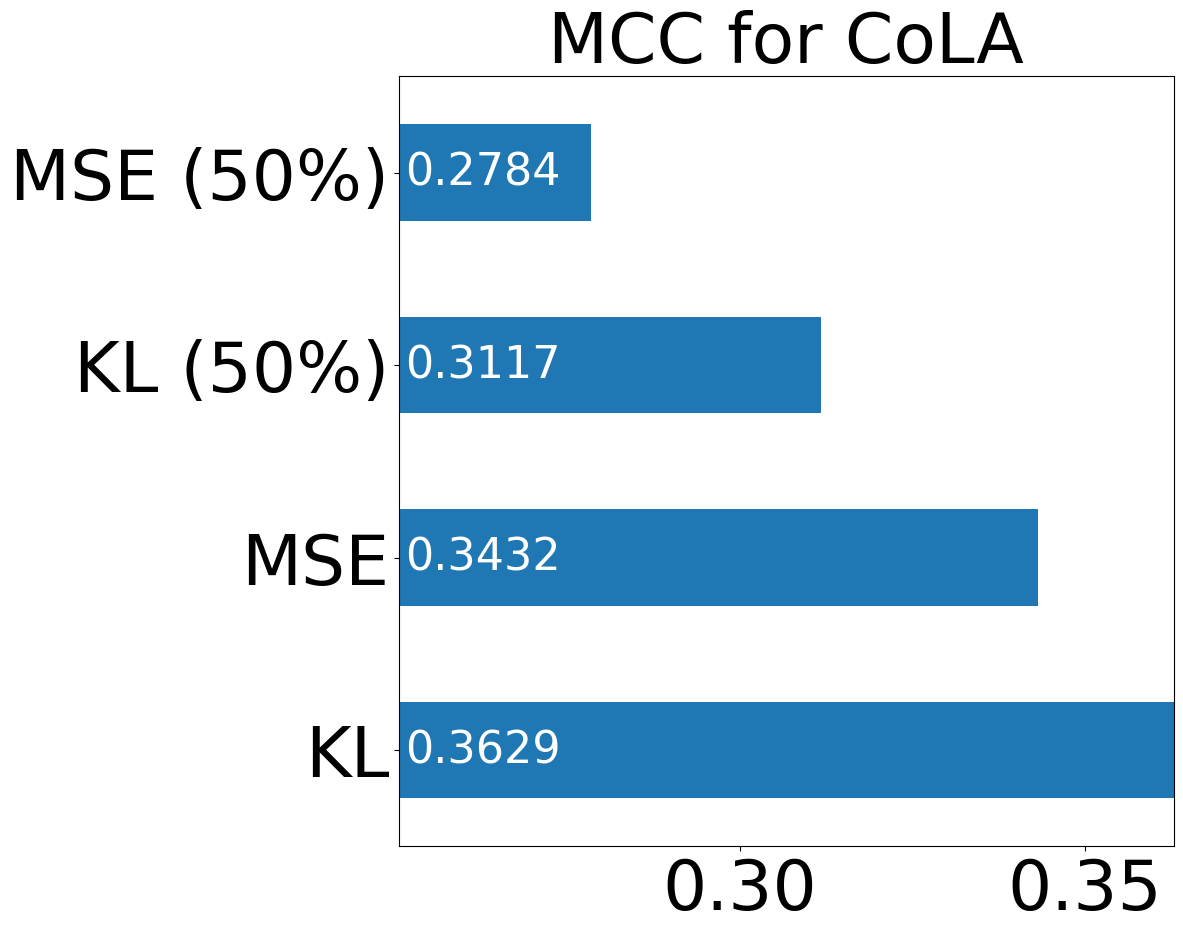} 
    \caption{Loss Function}
  \end{subfigure}
  \hfill
  \begin{subfigure}{0.28\textwidth}
    \includegraphics[width=\linewidth]{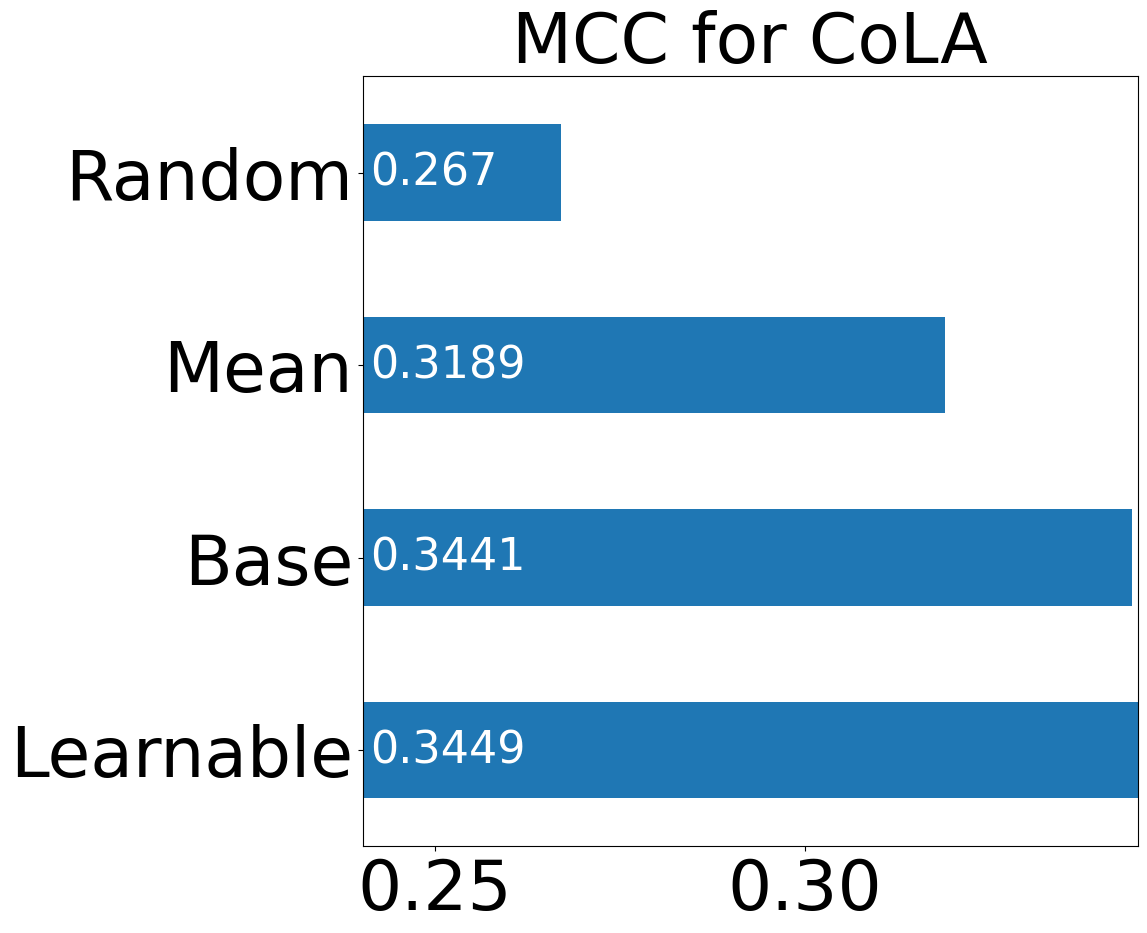} 
    \caption{Mapping Function}
  \end{subfigure}
  \hfill
  \begin{subfigure}{0.34\textwidth}
    \includegraphics[width=\linewidth]{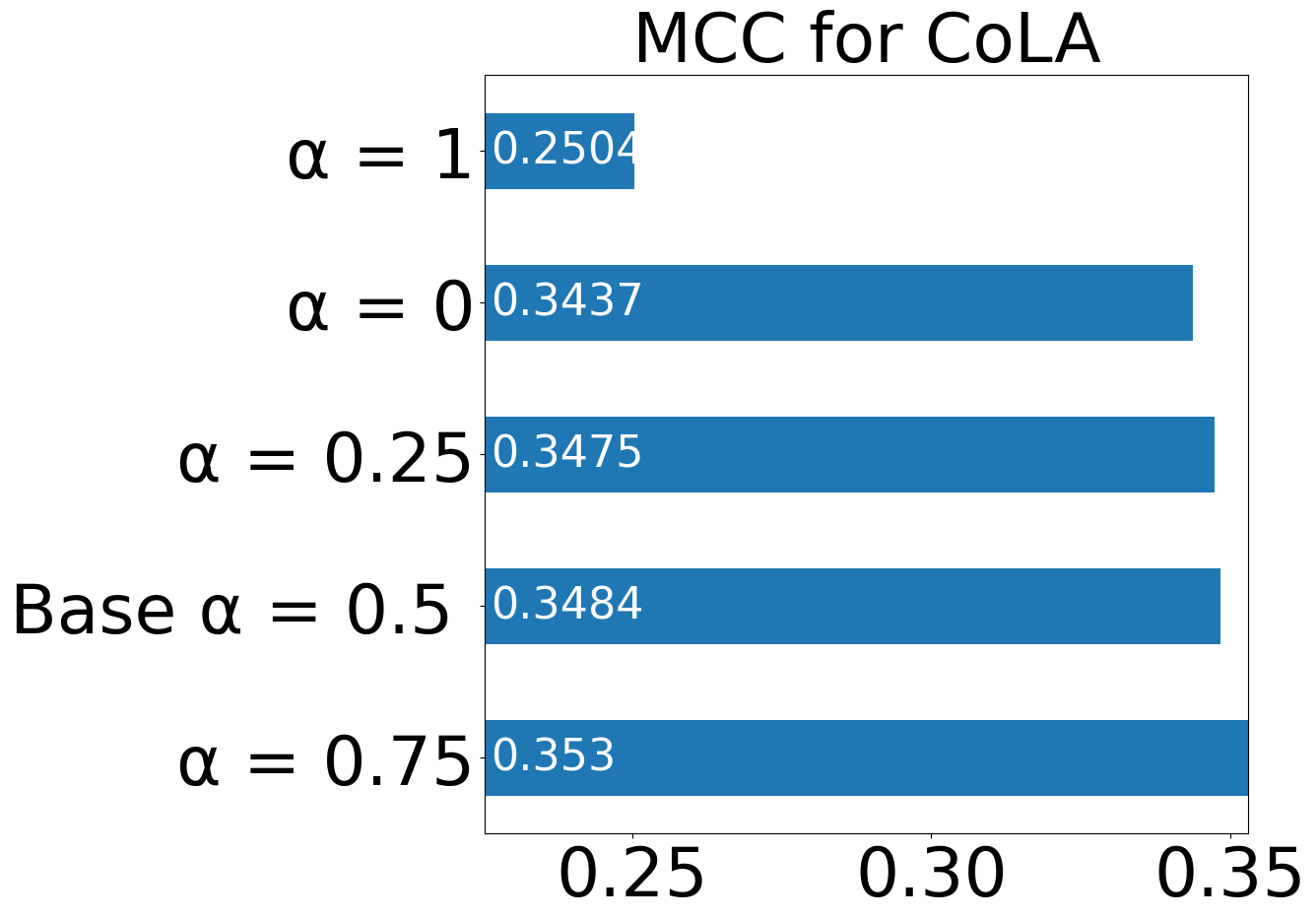} 
    \caption{Weights of Rep Loss}
  \end{subfigure}
  \caption{Various experiments for CoLA }
  \label{fig:cola_results}
\end{figure}

\begin{figure}[h]
  \centering
  \begin{subfigure}{0.28\textwidth}
    \includegraphics[width=\linewidth]{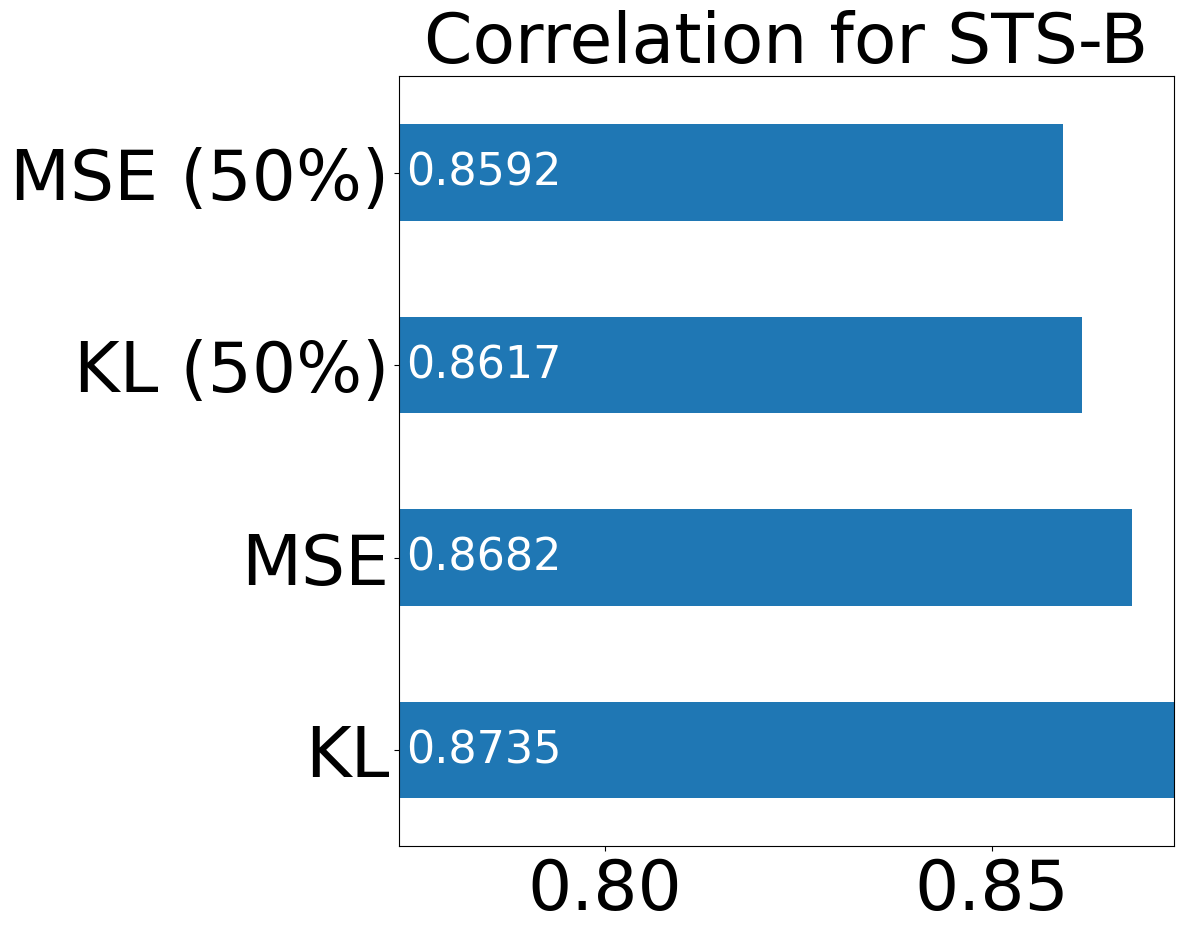} 
    \caption{Loss Function}
  \end{subfigure}
  \hfill
  \begin{subfigure}{0.28\textwidth}
    \includegraphics[width=\linewidth]{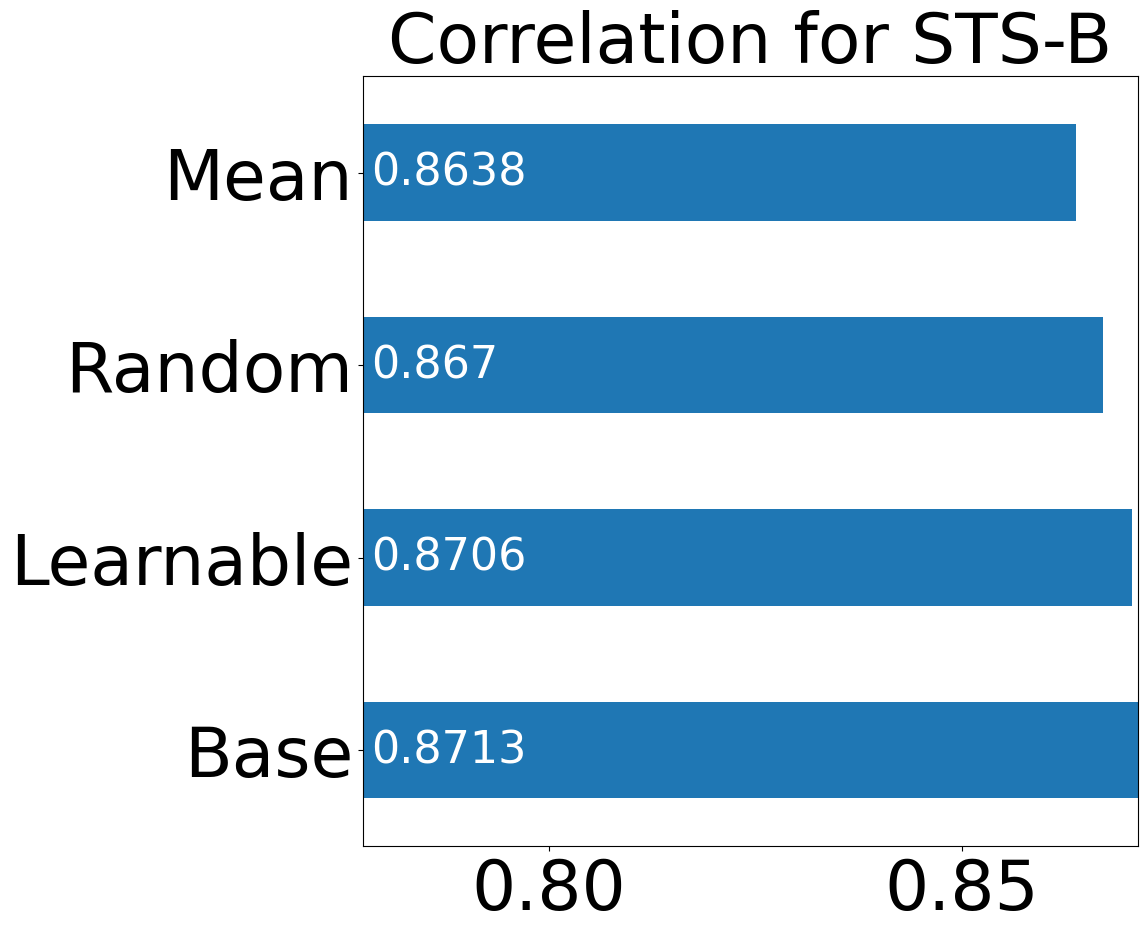} 
    \caption{Mapping Function}
  \end{subfigure}
  \hfill
  \begin{subfigure}{0.34\textwidth}
    \includegraphics[width=\linewidth]{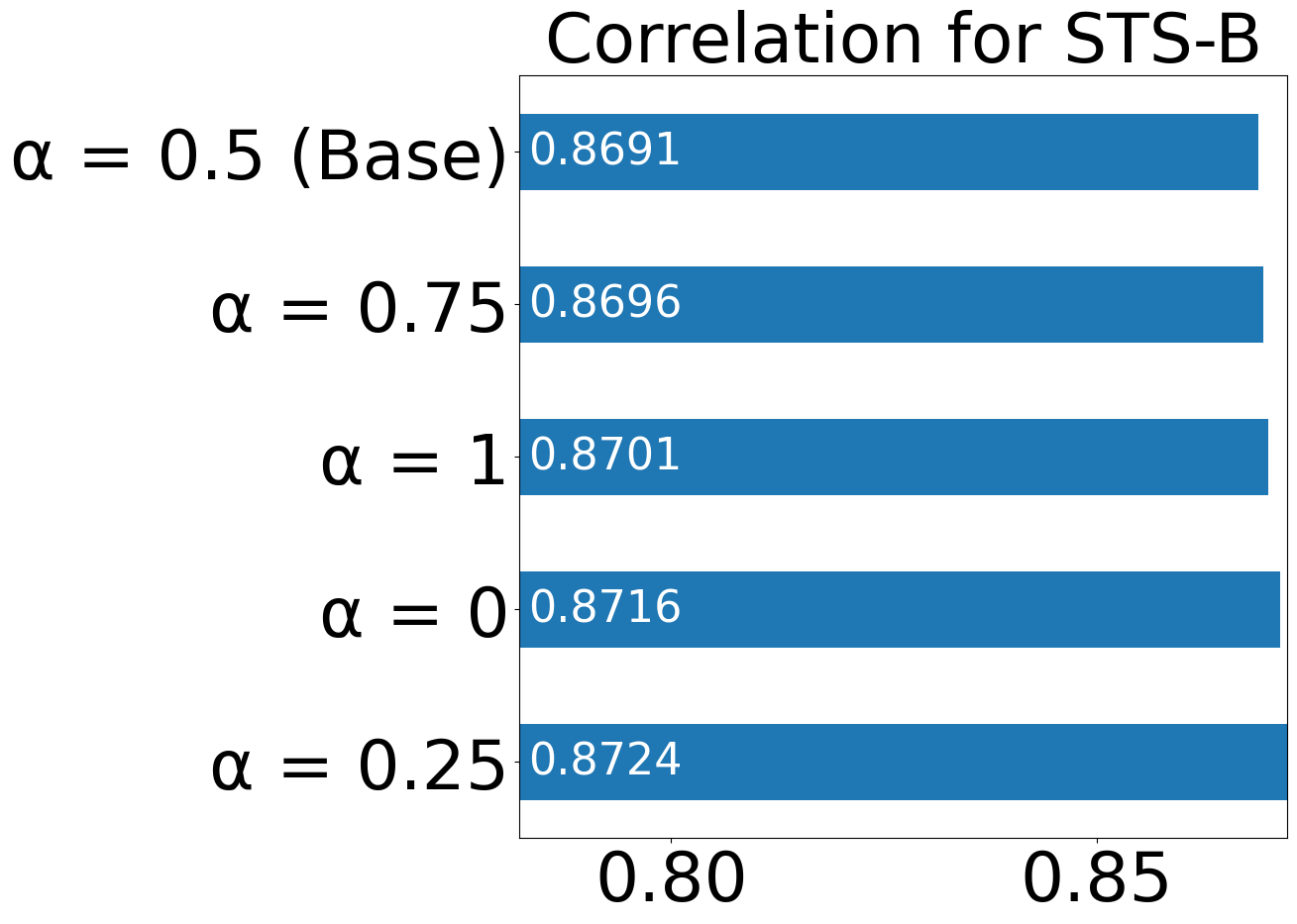} 
    \caption{Weights of Rep Loss}
  \end{subfigure}
  \caption{Various experiments for STS-B }
  \label{fig:stsb_results}
\end{figure}

In our experiments, we investigated the impact of different values of $\alpha$ on task-specific distillation performance. We observed that for the CoLA task, setting $\alpha = 1$ resulted in significantly worse performance compared to other values, while other values performed comparably. This finding suggests that attention loss plays a crucial role in the CoLA task. On the other hand, for the STS-B task, we found that all tested $\alpha$ values returned similar results (\ref{stsb_withouts1}). The detailed results can be found in the right panels of Figures \ref{fig:cola_results} and \ref{fig:stsb_results}. 

\section{Conclusion}

In this study, we investigated various factors influencing the performance of the TinyBERT model during task-specific distillation. Our findings indicate that the KL divergence loss function can offer performance improvements for challenging tasks like CoLA, especially when training with limited data. For mapping functions, while the learnable mapping didn't yield significant improvements, we observed that weight vectors converged to the same values regardless of initialization, hinting at a potentially interesting area for future exploration. Our experiments on weighting attention and representation loss revealed that attention loss plays a crucial role in the CoLA task, while having less impact on the STS-B task. 




In the case of the STS-B task, the performance did not change significantly in any experiment, indicating that the model may not be learning much from transformer loss and that most learning is happening through prediction loss distillation. We later confirmed this by skipping the transformer loss distillation and achieving similar results as when it was included (\ref{stsb_withouts1}).


It is important to note that since we experimented only with the task-specific distillation stage, the performance improvements observed may be further enhanced if we apply the same modifications in the general distillation stage as well. Overall, our research provides valuable insights for future work on improving transformer-based model distillation methods.

\bibliographystyle{plainnat}
\bibliography{main} 

\begin{thebibliography}{11}
\providecommand{\natexlab}[1]{#1}
\providecommand{\url}[1]{\texttt{#1}}
\expandafter\ifx\csname urlstyle\endcsname\relax
  \providecommand{\doi}[1]{doi: #1}\else
  \providecommand{\doi}{doi: \begingroup \urlstyle{rm}\Url}\fi

\bibitem[Biewald(2020)]{biewald2020experiment}
Lukas Biewald.
\newblock Experiment tracking with weights and biases, 2020.
\newblock \emph{Software available from wandb. com}, 2\penalty0 (5), 2020.

\bibitem[Cer et~al.(2017)Cer, Diab, Agirre, Lopez-Gazpio, and
  Specia]{cer2017semeval}
Daniel Cer, Mona Diab, Eneko Agirre, Inigo Lopez-Gazpio, and Lucia Specia.
\newblock Semeval-2017 task 1: Semantic textual similarity-multilingual and
  cross-lingual focused evaluation.
\newblock \emph{arXiv preprint arXiv:1708.00055}, 2017.

\bibitem[Devlin et~al.(2018)Devlin, Chang, Lee, and
  Toutanova]{devlin2018pretraining}
Jacob Devlin, Ming-Wei Chang, Kenton Lee, and Kristina Toutanova.
\newblock Bert: Pre-training of deep bidirectional transformers for language
  understanding, 2018.
\newblock URL \url{http://arxiv.org/abs/1810.04805}.
\newblock cite arxiv:1810.04805Comment: 13 pages.

\bibitem[Hinton et~al.(2015)Hinton, Vinyals, and Dean]{hinton2015distilling}
Geoffrey Hinton, Oriol Vinyals, and Jeff Dean.
\newblock Distilling the knowledge in a neural network, 2015.
\newblock URL \url{http://arxiv.org/abs/1503.02531}.
\newblock cite arxiv:1503.02531Comment: NIPS 2014 Deep Learning Workshop.

\bibitem[Jiao et~al.(2019)Jiao, Yin, Shang, Jiang, Chen, Li, Wang, and
  Liu]{jiao2019tinybert}
Xiaoqi Jiao, Yichun Yin, Lifeng Shang, Xin Jiang, Xiao Chen, Linlin Li, Fang
  Wang, and Qun Liu.
\newblock Tinybert: Distilling bert for natural language understanding.
\newblock \emph{arXiv preprint arXiv:1909.10351}, 2019.

\bibitem[Kullback and Leibler(1951)]{kullback1951information}
Solomon Kullback and Richard~A Leibler.
\newblock On information and sufficiency.
\newblock \emph{The annals of mathematical statistics}, 22\penalty0
  (1):\penalty0 79--86, 1951.

\bibitem[Morris et~al.(2020)Morris, Lifland, Yoo, Grigsby, Jin, and
  Qi]{morris2020textattack}
John~X. Morris, Eli Lifland, Jin~Yong Yoo, Jake Grigsby, Di~Jin, and Yanjun Qi.
\newblock Textattack: A framework for adversarial attacks, data augmentation,
  and adversarial training in nlp, 2020.

\bibitem[Sanh et~al.(2019)Sanh, Debut, Chaumond, and
  Wolf]{Sanh2019DistilBERTAD}
Victor Sanh, Lysandre Debut, Julien Chaumond, and Thomas Wolf.
\newblock Distilbert, a distilled version of bert: smaller, faster, cheaper and
  lighter.
\newblock \emph{ArXiv}, abs/1910.01108, 2019.

\bibitem[Sun et~al.(2019)Sun, Cheng, Gan, and Liu]{Sun2019PatientKD}
S.~Sun, Yu~Cheng, Zhe Gan, and Jingjing Liu.
\newblock Patient knowledge distillation for bert model compression.
\newblock In \emph{Conference on Empirical Methods in Natural Language
  Processing}, 2019.

\bibitem[Wang et~al.(2018)Wang, Singh, Michael, Hill, Levy, and
  Bowman]{wang-etal-2018-glue}
Alex Wang, Amanpreet Singh, Julian Michael, Felix Hill, Omer Levy, and Samuel
  Bowman.
\newblock {GLUE}: A multi-task benchmark and analysis platform for natural
  language understanding.
\newblock In \emph{Proceedings of the 2018 {EMNLP} Workshop {B}lackbox{NLP}:
  Analyzing and Interpreting Neural Networks for {NLP}}, pages 353--355,
  Brussels, Belgium, November 2018. Association for Computational Linguistics.
\newblock \doi{10.18653/v1/W18-5446}.
\newblock URL \url{https://aclanthology.org/W18-5446}.

\bibitem[Warstadt et~al.(2018)Warstadt, Singh, and Bowman]{warstadt2018neural}
Alex Warstadt, Amanpreet Singh, and Samuel~R Bowman.
\newblock Neural network acceptability judgments.
\newblock \emph{arXiv preprint arXiv:1805.12471}, 2018.

\end{thebibliography}
\vspace{25pt}
\section*{Team Contribution}

Authors are listed in alphabetical order. All members contributed equally throughout the duration of the project. All decisions related to the different parts of the project (like problem formulation, literature review, experimental design, and inferences) were taken collectively after discussion. Some specific contributions are as follows: Adeem took the initiative to implement the loss functions and run experiments. Apoorv did the same with various Mapping functions. Kartikaeya was responsible for implementing and experimenting with weight tuning.

\newpage
\appendix
\section{Appendix}

\subsection{Converging weights from LearnableMap}
\label{weights_learnablemap}

One of the very interesting things that we observed was that while using LearnableMap, with the correct learning rate and number of epochs we converge on the same final weights of the learnable parameter $\theta$. This finding indicates that there exists an optimal ${v}(m)$ that minimizes the transformer loss. However, we gathered this insight late in the project and could not experiment with it enough to get the best results from this. The following plot shows the convergence and optimal value of weights. Here $v(m)[j]$ denotes the value of the learned weight for $j^{th}$ layer in the respective block for the $m^{th}$ student layer.

\begin{figure}[h]
\centering
\includegraphics[scale=0.35]{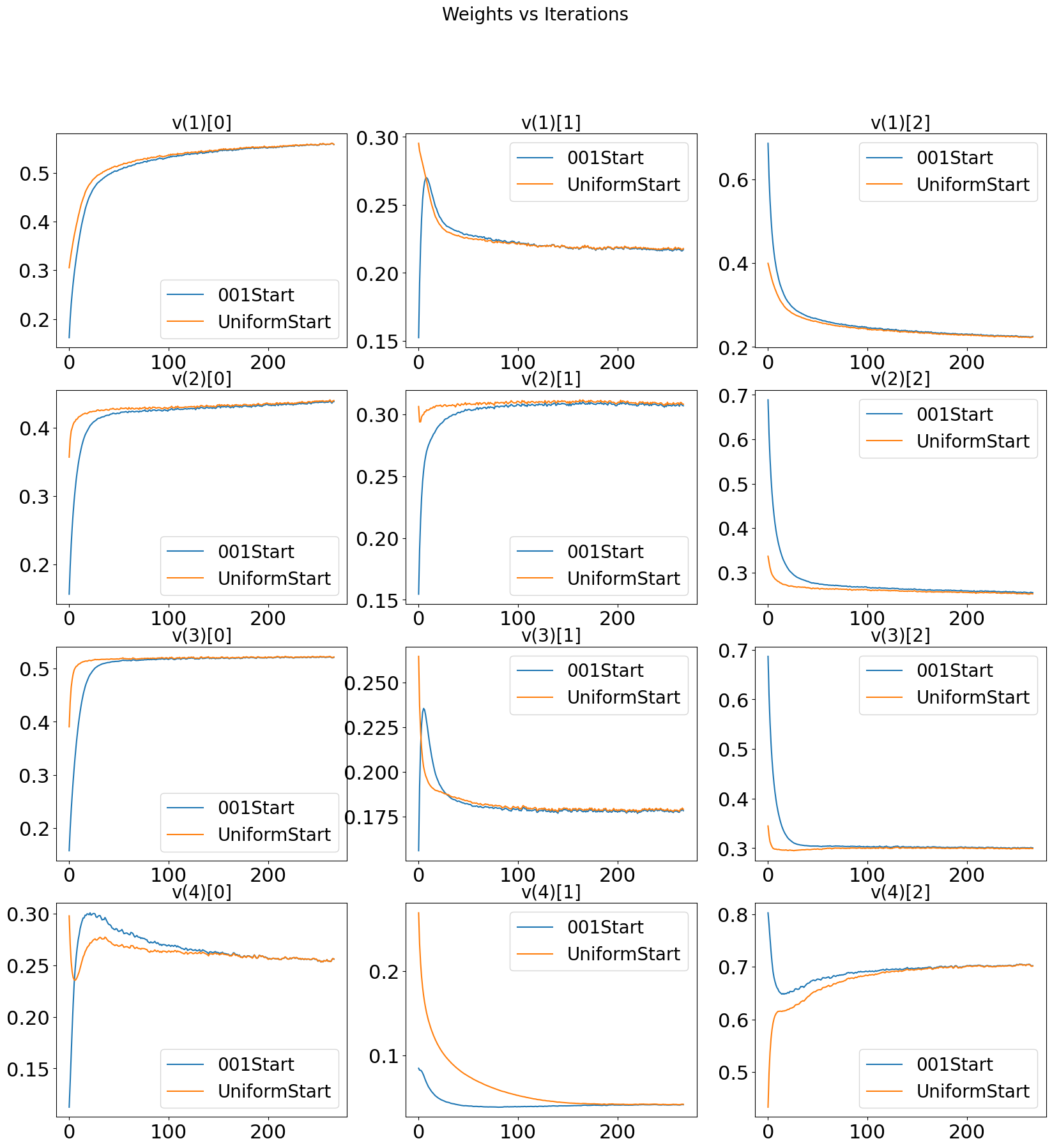}
\caption{Converging weights from LearnableMap}
\label{fig:converging_weight} 
\end{figure}

\subsection{Details for reproducibility}
\label{reproducibility}

In the following tables, we have reported what hyperparameter combinations gave us the best performance metrics for the experiments. The learning rates and batch size are given for the prediction distillation substage. 

For the KL loss, we experimented with different weights relative to the representation loss. We observed that a weight of 1 resulted in good performance, and increasing the weight beyond that led to only marginal improvements.

GitHub Links:
\begin{itemize}
    \item \href{https://github.com/apoorvdankar/Pretrained-Language-Model/tree/adeem/TinyBERT}{\textcolor{blue}{Loss Function Experiment}} 
    \item \href{https://github.com/apoorvdankar/Pretrained-Language-Model/tree/apoorv/TinyBERT}{\textcolor{blue}{Mapping Function Experiment}}
    \item \href{https://github.com/apoorvdankar/Pretrained-Language-Model/tree/kartikaeya/TinyBERT}{\textcolor{blue}{Weight Tuning Experiment}}
\end{itemize}

Other hyperparameters values:
\begin{itemize}
    \item Random Seed is set at 42.
    \item Learning rate for the transformer layer stage is 5e-05 
    \item Stage 1 epochs for CoLA is 30
    \item Stage 1 epochs for STS-B is 20
    \item Stage 2 epochs for both is 3
    \item Stage 1 batch size is 32
    \item For Mapping Function experiments the run\_name args should be one of the following (or contain these strings in the name) : [ RandomMap, MeanMap, LearnableMap ]
    \item To specify an initialization of the Learnable map, these run\_name should be used : [ LearnableMapUniformStart, LearnableMap001Start ]
\end{itemize}

\begin{table}[!hp]
\centering
\begin{tabular}{|l|c|c|c|}
\hline
\textbf{Loss Function} & \textbf{MCC} & \textbf{Learning Rate} & \textbf{Batch Size} \\
\hline
KL & 0.362875 & 1e-05 & 16 \\
MSE & 0.343187 & 1e-05 & 16 \\
KL (50\%) & 0.311703 & 2e-05 & 32 \\
MSE (50\%) & 0.278384 & 2e-05 & 16 \\
\hline
\end{tabular}
\caption{Best hyperparameters for Loss function experiments on CoLA}
\end{table}

\begin{table}[!hp]
\centering
\begin{tabular}{|l|c|c|c|}
\hline
\textbf{Loss Function} & \textbf{Corr} & \textbf{Learning Rate} & \textbf{Batch Size} \\
\hline
KL & 0.873543 & 3e-05 & 16 \\
MSE & 0.868203 & e-05 & 32 \\
KL (50\%) & 0.861659 & 3e-05 & 16 \\
MSE (50\%) & 0.859205 & 3e-05 & 16 \\
\hline
\end{tabular}
\caption{Best hyperparameters for Loss function experiments on STS-B}
\end{table}



\begin{table}[!hp]
\centering
\begin{tabular}{|l|c|c|c|}
\hline
\textbf{Mapping Function} & \textbf{MCC} & \textbf{Learning Rate (LearnableMap lr)} & \textbf{Batch Size} \\
\hline
Learnable & 0.344937 & 1e-05 (5e-05) & 32 \\
Base & 0.344144 & 2e-05 & 16 \\
Mean & 0.318876 & 1e-05 & 32 \\
Random & 0.266996 & 1e-05 & 32\\
\hline
\end{tabular}
\caption{Best hyperparameters for Mapping Function experiments on CoLA}
\end{table}

\begin{table}[!hp]
\centering
\begin{tabular}{|l|c|c|c|}
\hline
\textbf{Mapping Function} & \textbf{Corr} & \textbf{Learning Rate (LearnableMap lr)} & \textbf{Batch Size} \\
\hline
Base & 0.871267 & 3e-05 & 16\\
Learnable & 0.870572 & 3e-05 (5e-05) & 32\\
Random & 0.867048 & 3e-05 & 16\\
Mean & 0.863824 & 3e-05 & 16 \\
\hline
\end{tabular}
\caption{Best hyperparameters for Mapping Function experiments on STS-B}
\end{table}

\begin{table}[!hp]
\centering
\begin{tabular}{|l|c|c|c|}
\hline
\textbf{Weight of Rep Loss} & \textbf{MCC} & \textbf{Learning Rate} & \textbf{Batch Size} \\
\hline
$\alpha = 0.75$ & 0.353044 & 2e-05 & 32 \\
$\alpha = 0.5$ (Base) & 0.348409 & 2e-05 & 32 \\
$\alpha = 0.25$ & 0.347475 & 1e-05 & 32 \\
$\alpha = 0$ & 0.343706 & 3e-05 & 32 \\
$\alpha = 1$ & 0.250387 & 3e-05 & 32 \\
\hline
\end{tabular}
\caption{Best hyperparameters for $\alpha$ tuning experiments on CoLA}
\end{table}

\begin{table}[!hp]
\centering
\begin{tabular}{|l|c|c|c|}
\hline
\textbf{Weight of Rep Loss} & \textbf{Corr} & \textbf{Learning Rate} & \textbf{Batch Size} \\
\hline
$\alpha = 0.25$ & 0.872409 & 3e-05 & 32 \\
$\alpha = 0$ & 0.871626 & 3e-05 & 16\\
$\alpha = 1$ & 0.870140 & 2e-05 & 16\\
$\alpha = 0.75$ & 0.869641 & 3e-05 & 32 \\
$\alpha = 0.5$ (Base) & 0.869064 & 2e-05 & 32\\
\hline
\end{tabular}
\caption{Best hyperparameters for $\alpha$ tuning experiments on STS-B}
\end{table}

\subsection{Confirmation of the Irrelevance of Transformer Layer Loss Distillation for STS-B Task}
\label{stsb_withouts1}
In our previous experiments, we observed no significant change in the results of the STS-B task, leading us to hypothesize that transformer layer distillation might not be crucial for this task. To test this hypothesis, we conducted an experiment in which we skipped the transformer layer loss distillation step.

Instead of fine-tuning the TinyBERT model using the transformer layer loss, we directly trained the model on the prediction loss distillation task, using the general TinyBERT model as the initialization. Interestingly, we found that skipping the transformer layer loss distillation stage yielded almost the same result (corr = 0.8717) as the baseline (corr = 0.8682). In fact, the performance was slightly better when skipping the transformer layer loss distillation.

This observation suggests that the transformer layer loss distillation might not be essential for the STS-B task, and a more efficient alternative could involve directly training on the prediction loss distillation task. Additionally, this experiment highlights the possibility that the relevance of transformer layer loss distillation may depend on the specific task, and it might not be equally important for all tasks.


\end{document}